\documentclass[a4paper,12pt]{article}
\usepackage[a4paper,left = 2cm, right= 2cm,top = 2cm, bottom = 2cm]{geometry}
\usepackage{cite}
\usepackage{amsmath,amssymb,amsfonts}
\usepackage{multirow}
\usepackage{algorithmic}
\usepackage{graphicx}
\usepackage{textcomp}
\usepackage{amssymb,
amsthm,
amsfonts,
amsmath,
enumerate,
caption,
subfigure,diagbox,booktabs
}
\usepackage[ruled,vlined,commentsnumbered]{algorithm2e}

\usepackage{authblk}

\begin{document}
\bibliographystyle{plain}
\title{HB-net: Holistic bursting cell cluster integrated network for occluded multi-objects recognition}
\date{}
\author[a]{Xudong Gao} 
\author[a]{Xiaoguang Gao} 
\author[b]{Jia Rong} 
\author[c,d]{Xiaowei Chen} 
\author[e]{Xiang Liao\thanks{Corresponding Author, e-mail:xiang.liao@cqu.edu.cn}}
\author[a,d]{Jun Chen\thanks{Corresponding Author, e-mail:junchen@nwpu.edu.cn}}

\affil[a]{School of Electronics and Information, Northwestern Polytechnical University, Xi'an 710072, Shaanxi, China}
\affil[b]{Department of Data Science and AI, Monash University, Clayton, Melbourne,VIC3800, Victoria, Australia}
\affil[c]{Brain Research Center and State Key Laboratory of Trauma, Burns, and Combined Injury, Third Military Medical University, 400038, Chongqing, China}
\affil[d]{Chongqing Institute for Brain and Intelligence, Guangyang Bay Laboratory, Nan'an District, 400064, Chongqing, China}
\affil[e]{Center for Neurointelligence,  School of Medcine, Chongqing University, 131 Yubei Road, Shapingba District, 400030, Chongqing, China}

\newcommand\blfootnote[1]{%
\begingroup
\renewcommand\thefootnote{}\footnote{#1}%
\addtocounter{footnote}{-1}%
\endgroup
}

\maketitle

\begin{abstract}
  Within the realm of image recognition, a specific category of multi-label classification (MLC) challenges arises when objects within the visual field may occlude one another, demanding simultaneous identification of both occluded and occluding objects. 
  Traditional convolutional neural networks (CNNs) can tackle these challenges; however, those models tend to be bulky and can only attain modest levels of accuracy. 
  Leveraging insights from cutting-edge neural science research, specifically the Holistic Bursting (HB) cell, this paper introduces a pioneering integrated network framework named HB-net. 
  Built upon the foundation of HB cell clusters, HB-net is designed to address the intricate task of simultaneously recognizing multiple occluded objects within images.
  Various Bursting cell cluster structures are introduced, complemented by an evidence accumulation mechanism. 
  Testing is conducted on multiple datasets comprising digits and letters. 
  The results demonstrate that models incorporating the HB framework exhibit a significant $2.98\%$ enhancement in recognition accuracy compared to models without the HB framework ($1.0298$ times, $p=0.0499$). 
  Although in high-noise settings, standard CNNs exhibit slightly greater robustness when compared to HB-net models,
  the models that combine the HB framework and EA mechanism achieve a comparable level of accuracy and resilience to ResNet50, despite having only three convolutional layers and approximately $1/30$ of the parameters. 
  The findings of this study offer valuable insights for improving computer vision algorithms. 
  The essential code is provided at https://github.com/d-lab438/hb-net.git.

\end{abstract}

\section{Introduction}  \label{sec:introduction}

In natural visual environments, occlusion frequently occurs when objects of interest are either wholly or partially obscured by other objects. 
Disregarding these occluding objects is inadequate, as they may also bear significance for the observer.
Consequently, the recognition of occluded objects within images becomes a multi-label classification (MLC) challenge.
In such scenarios, objects of interest are partially or entirely concealed by other elements, demanding that the classifier proficiently discern both the obscured objects and those causing the occlusion. 
The classifier must not only identify the shape and texture characteristics of the occluded objects but also leverage the partly visible shape and texture attributes to recognize the obscured objects.
Undeniably, this significantly heightens the complexity of occluded object recognition.

The challenge of MLC under occlusion conditions holds significant importance across diverse fields and practical applications.
In medical image analysis for healthcare \cite{Cui2022}, encompassing Magnetic Resonance Imaging (MRI), X-rays, and Computed Tomography (CT) scans, the accurate identification and classification of diseases, lesions, and anatomical structures become notably complex due to occlusion and incomplete image data. 
Addressing the MLC under occlusion conditions empowers doctors and medical professionals to enhance their understanding and diagnostic accuracy of medical images.
In the human-computer interaction domain, MLC under occlusion conditions is pivotal for improving human-computer interaction \cite{Kao2019, Sheng2023}. 
For instance, in face recognition \cite{Lokku2022} and emotion analysis \cite{Yuan2021}, occluded facial features can lead to recognition errors and inaccurate emotion analysis. 
Solving the challenge of MLC under occlusion conditions elevates the accuracy and overall interaction experience in human-computer interfaces.
Occlusion frequently poses challenges in gesture recognition and action analysis, especially when hand or body parts are obscured by other objects \cite{Liao2021}. 
Tackling the MLC under occlusion conditions bolsters the performance of gesture recognition systems, facilitating more precise and dependable gesture-based interactions.
Within the agricultural domain, multi-label classification under occlusion conditions finds utility in crop identification and disease/pest detection \cite{Wang2022}. 
Resolving this issue empowers farmers and agricultural professionals to accurately identify and classify crops and disease/pest infestations, enabling timely interventions to safeguard crops and enhance agricultural productivity.

\begin{figure}[htbp]
  \centering
    \includegraphics[width=0.45\textwidth]{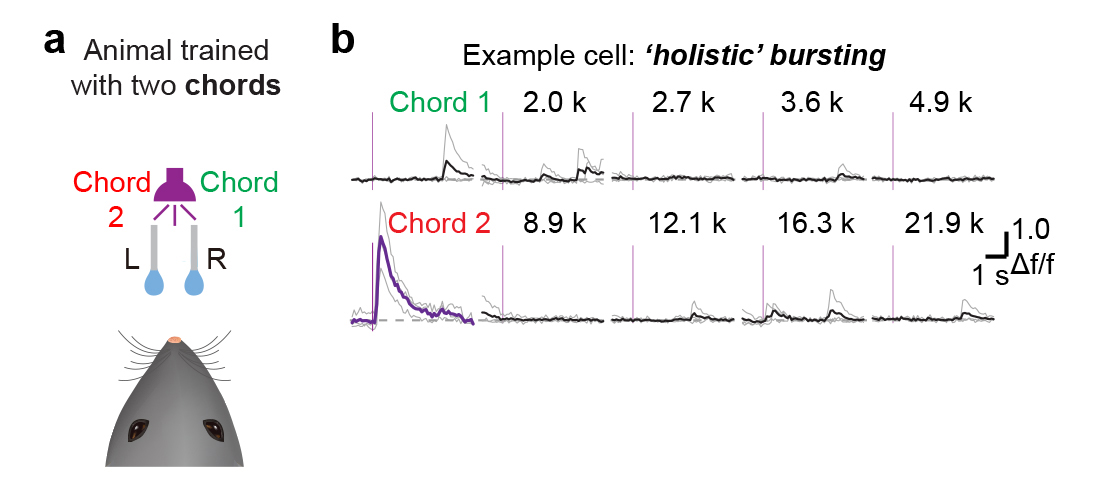}
  \caption{The HB cell in the brain of mouse \cite{Wang2020}.}
  \label{Real_HB} 
\end{figure}

Neuroscience and artificial intelligence share a symbiotic relationship, with progress in one often sparking advances in the other.
In $2021$, XiaoWei Chen's team employed a blend of two-photon calcium imaging and single-cell electrophysiology to investigate the spike firing patterns of individual cortical neurons in awake mice following auditory associative training \cite{Wang2020}, depicted in Fig. \ref{Real_HB}. 
Their research revealed that approximately $5\%$ of primary auditory cortex neurons exhibited high-frequency, long-pulse responses to the trained sound, a rarity in untrained mice.
Furthermore, during training involving different polyphonic chords, the researchers identified specific neuron subsets responsive solely to one chord's pulse firing, excluding other components or chords. 
This finding implies that mice activate distinct cells when recognizing specific sounds, which they termed Holistic Bursting (HB) cells. 
This concept serves as an inspiration for using dedicated networks to recognize specific objects in multi-object recognition, culminating in a final result through a merging network.
This approach introduces a novel method to tackle the challenge of accurately identifying all objects in scenarios where multiple objects within the same field of view are occluded. 
It involves creating a network composed of several sub-networks, each serving as a virtual HB cell cluster responsible for recognizing specific object categories within the image and learning their unique characteristics. 
Consequently, the outputs from all sub-networks are combined to yield the ultimate recognition results. 
Therefore, this paper presents the design of an integration network model framework named "HB-net," founded on clusters of HB cells, to address the issue of multi-object occlusion within a single image.

The primary contributions and innovations of this paper can be summarized as follows:

\begin{itemize}
  \item Leveraging recent advancements in computational neuroscience, we devised the HB-net framework, which emulates brain computation principles. 
  This modular and integrated structure was designed to address the challenge of multi-object occlusion in images and has demonstrated notable success. 
  This innovative approach offers a fresh technical perspective and holds substantial bio-cognitive inspiration.
  \item We introduced various holistic burst cell cluster structures, including bottom-up (B), bottom-up with lateral (BL), bottom-up with top-down (BT), and bottom-up with both lateral and top-down (BLT). 
  Additionally, we incorporated the evidence accumulation (EA) mechanism, expanding the model's architecture and enhancing its structural diversity. 
  This innovation broadens the model's design possibilities and increases its adaptability.
  \item Comprehensive testing on various models across multiple datasets was conducted to validate the effectiveness of the HB framework and the evidence accumulation mechanism. 
  Additionally, in-depth analyses of model robustness in noisy environments were performed. 
  These empirical results furnish substantial evidence in support of our approach.
\end{itemize}
These contributions and innovations endow this research with significant value and significance in the field of computer vision.

The rest of the paper is organized as follows: Section \ref{related_work} reviews the methodologies and achievements in the recent work of occluded multi-object recognition. Section \ref{method} discusses the datasets used and the construction of models. Section \ref{experiment} outlines the experimental design. Section \ref{results} presents the experimental results. Section \ref{discussion} engages in a discussion of the relevant findings. Section \ref{conclusion} provides conclusions, followed by a consideration of potential future research directions.

\section{Related work} \label{related_work}

Scholars have undertaken extensive research to tackle the challenge of object occlusion. 
The methods employed typically encompass Convolutional Neural Networks (CNN), Recurrent Convolutional Neural Networks (RCNN), Principal Component Analysis (PCA), Independent Component Analysis (ICA), dictionary construction, attention mechanisms, autoencoders, generative adversarial networks, and various other techniques.

When using Convolutional Neural Network (CNN) models to extract occluded facial features, the occluded parts are embedded in the representation \cite{Georgescu2022}. 
One approach to address this is to synthesize occluded parts to expand the dataset and subsequently perform recognition, while another approach entails segmenting the occluded parts before conducting recognition. 
Sarapakdi and Sittiphan \cite{Sarapakdi2019} investigated the impact of Two-Dimensional Principal Component Analysis (2DPCA) on convolutional neural network image classification, resulting in reduced complexity and enhanced accuracy for facial occlusion recognition systems. 
Kortylewski and Adam \cite{Kortylewski2020,Kortylewski2021} improved face recognition accuracy through three-dimensional face reconstruction and sequential deep learning, showcasing robustness against occlusion and facial expressions. 
Soni and Neha \cite{Soni2021} implemented a simple and efficient facial recognition system for handling occlusion and noise, achieving strong performance using deep learning. 
Later, a Upper and Lower Branch Network (ULN) architecture \cite{Georgescu2022} was proposed and aimed at efficient identification of masked faces. 
Furthermore, these methods find application in a variety of contexts, including object detection in classroom settings \cite{Ding2020}, tomato ripeness and occlusion degree detection \cite{Wang2022}, occlusion modeling for crowd counting \cite{Almalki2021, Zhu2018}, improvements in network models for occluded gesture recognition \cite{Liao2021}, occlusion-aware re-identification (Re-ID) methods \cite{Li2022}, and beyond.

An alternative approach is to use Recurrent Convolutional Neural Network (RCNN) to handle occluded noise. 
The temporal dynamics of object recognition under occlusion conditions using magnetoencephalography (MEG) filled a knowledge gap and shedded light on the mechanisms of object recognition under occlusion \cite{Rajaei2019}. 
This model revealed the impact of occlusion on the neural dynamics of object recognition. 
Spoerer et al. introduced a recurrent neural network with bottom-up, lateral, and top-down connections to simulate the neural dynamics of the ventral visual pathway, and using evidence accumulation (EA) mechanism, thereby improving performance in object recognition under occlusion \cite{Spoerer2017,Spoerer2020,Kietzmann2019}. 

Dictionary construction (DC) is a prevalent technique for managing occlusion challenges, with various contributions in this area \cite{Madarkar2020,Deng2020,Wen2016}. 
Yuan et al. \cite{Yuan2016} presented a non-negative dictionary method aimed at resolving occlusion-related issues in face recognition. Du et al. introduced a nuclear norm adaptive occlusion dictionary learning framework, enhancing the robustness of face recognition in the presence of occlusion \cite{Du2019}.

Attention mechanisms have shown promise in addressing occlusion issues with several noteworthy studies \cite{Ran2021,Zou2020,Ding2020a,Peng2019,Li2021a} including a dual-branch bidirectional attention module for extracting features from occluded faces \cite{Shakeel2022}, a convolutional neural network model based on the biological visual attention mechanism to effectively handles occluded images \cite{Xiao2021}, and a transformer-based method for facial expression recognition that utilizes attention mechanisms to capture important facial features \cite{Liu2023,Gao2021}.

Autoencoders are another technique for handling occlusion \cite{Abbas2021,Xudong2019}. Wang et al. \cite{Wang2019} proposed generalizable autoencoders and stacked generalizable autoencoders for occluded region inpainting and feature learning. Sun et al. proposed an end-to-end Autoencoder model for handling occluded images \cite{Sun2022}.

Generative Adversarial Networks (GANs) have also achieved some success in addressing occlusion problems. In this case, researchers improved face recognition accuracy by addressing occlusion removal and face frontalization \cite{Zhu2021,Li2020,Duan2021}. 

In addition to the methods mentioned earlier, there are other approaches for effectively managing occlusion challenges.
Yang et al. \cite{Yang2019} improved pedestrian re-identification performance under occlusion by utilizing Long Short-Term Memory (LSTM) to learn spatial dependencies. Zheng et al. proposed a facial expression recognition method based on multi-source feature fusion, which can handle occlusion and multiple expressions \cite{Zheng2020}. 

The studies mentioned above reveal that the majority of research in this domain primarily addresses occlusion within images, whether by removing occlusions, guiding neural networks to focus on unoccluded features, or generating the original images of occluded parts. 
However, there is a noticeable lack of research focused on tackling mutual occlusion among multiple objects within the same field of view. 
Notably, RCNN-based methods emerge as effective tools for handling the mutual occlusion of multiple labels. 
In light of these findings, this paper aims to leverage recent insights from neuroscience, particularly the HB cell structure, in conjunction with RCNN, to construct the HB-net neural network framework. 
This framework is engineered to closely emulate visual recognition mechanisms akin to those in primates. 
Extensive testing across diverse datasets and noise conditions is conducted to rigorously evaluate its performance.

\section{Methods and Model} \label{method}

This section presents the three datasets to be tested, followed by an in-depth exploration of the framework and model architecture of HB-net and the intricacies of the model training process.

\subsection{The Generated Datasets}

The datasets utilized in this study are generated using the program provided by Spoerer's team \cite{Spoerer2017}. 
These datasets consist of various digits or letters, all rendered in the same font, color, and size. 
The sole difference lies in the character's position, which is uniformly distributed at random. 
To heighten the challenge, we have intentionally manipulated the level of occlusion. 
The objective is to create demanding occlusion scenarios that facilitate a clear distinction between different models and allow for a comprehensive assessment of their performance.

The chosen experimental design systematically isolates and investigates the distinct influence of occlusion on object recognition. 
By meticulously controlling for other variables and prioritizing occlusion as the primary variable of interest, the precise mechanisms are delved through which occlusion affects recognition model performance. 
This approach helps to gain valuable insights into the impact of occlusion and paves the way for exploring potential solutions to enhance the resilience and accuracy of recognition models when confronted with occlusion challenges.

\begin{figure}[htbp]
  \centering
  \includegraphics[width=0.45\textwidth]{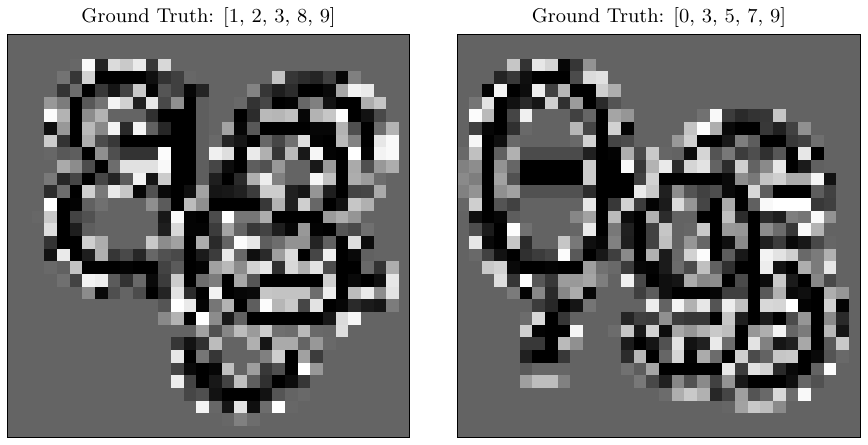}
  \caption{The examples of 5-digit clutter dataset.}
  \label{example_5} 
\end{figure}

In the experiments, a $5$-digit clutter dataset was employed, where five digits were sequentially generated and placed within an image. 
These digits' positions were randomly determined, adhering to a uniform distribution, resulting in overlapping digits and establishing a relative depth order. 
The network's objective was to recognize all the digits present in each image. 
Initially, these images were designed at a high resolution of $512 \times 512$ pixels. However, for the sake of computational efficiency, we resized them to a lower resolution of $32 \times 32$ pixels before input to the network. 
This resizing step strikes a balance between computational complexity and information sufficiency for the network to effectively address the task of digit recognition in occluded scenes. 
To ensure consistent and meaningful training, we generated random training and validation sets for each image collection. 
The training set comprised $100,000$ images, while the validation set included $10,000$ images, facilitating the determination of hyperparameters and learning regimes for our models. 
To evaluate model performance, an independent test set consisting of $10,000$ images was used. Examples of the $5$-digit clutter dataset are illustrated in Fig. \ref{example_5}.

\begin{figure}[htbp]
  \centering
  \includegraphics[width=0.45\textwidth]{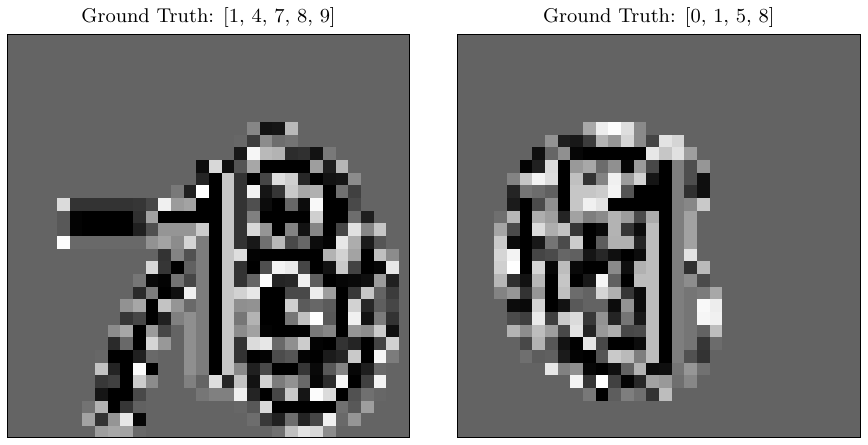}
  \caption{The examples of mixed 5-digit clutter dataset.}
  \label{example_mix} 
\end{figure}

To assess the models' adaptability to datasets with varying label counts, a mixed-digit clutter dataset was specifically generated, where each sample may encompass one or more digits, up to a maximum of five digits ($n \leq 5$). 
The corresponding training set contained an equal number of samples featuring different quantities of digits, with each number of digits represented by 20,000 samples. 
Consequently, the entire training set comprised $100,000$ samples. 
Similarly, the test set maintained an equal distribution of samples with different numbers of digits, with each category encompassing $2,000$ samples. 
The test set, thus, encompassed $10,000$ samples. Illustrated in Fig. \ref{example_mix} are examples from the mixed $5$-digit clutter dataset.

\begin{figure}[htbp]
  \centering
  \includegraphics[width=0.45\textwidth]{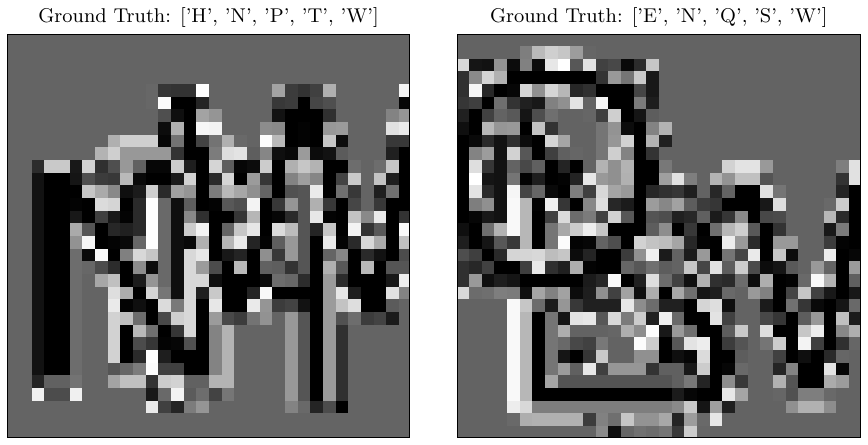}
  \caption{The examples of 5-letter clutter dataset.}
  \label{example_5character} 
\end{figure}

To assess the models' generalization to different datasets, another dataset was generated, where five English letters are occluded like $5$-digit clutter, called $5$-letter clutter dataset, as shown in Fig. \ref{example_5character}.

Prior to inputting the images into the network, a pixel-wise normalization was conducted. 
For each pixel $x$ at position $(i, j)$ within an image, the normalized pixel value $\widehat{x}_{i j}$ can be calculated using the following formula:
\begin{equation}
  \widehat{x}_{i j}=\frac{x_{i j}-\overline{{{x}}}_{i j}}{s_{x_{i j}}} 
\end{equation}
Here, $x_{i j}$ represents the raw pixel value, $\overline{{{x}}}_{i j}$ denotes the mean pixel value and $s_{x_{i j}} $represents the standard deviation of pixel values at that position. 
The mean and standard deviation values were computed for each specific position across the entire training dataset \cite{Spoerer2017}. 
This normalization step ensures that the pixel values are centered and scaled appropriately, aiding in effective training and comparison across different positions in the images.

\subsection{HB-net Framework}

\begin{figure}[htbp]
  \centering
    \includegraphics[width=0.45\textwidth]{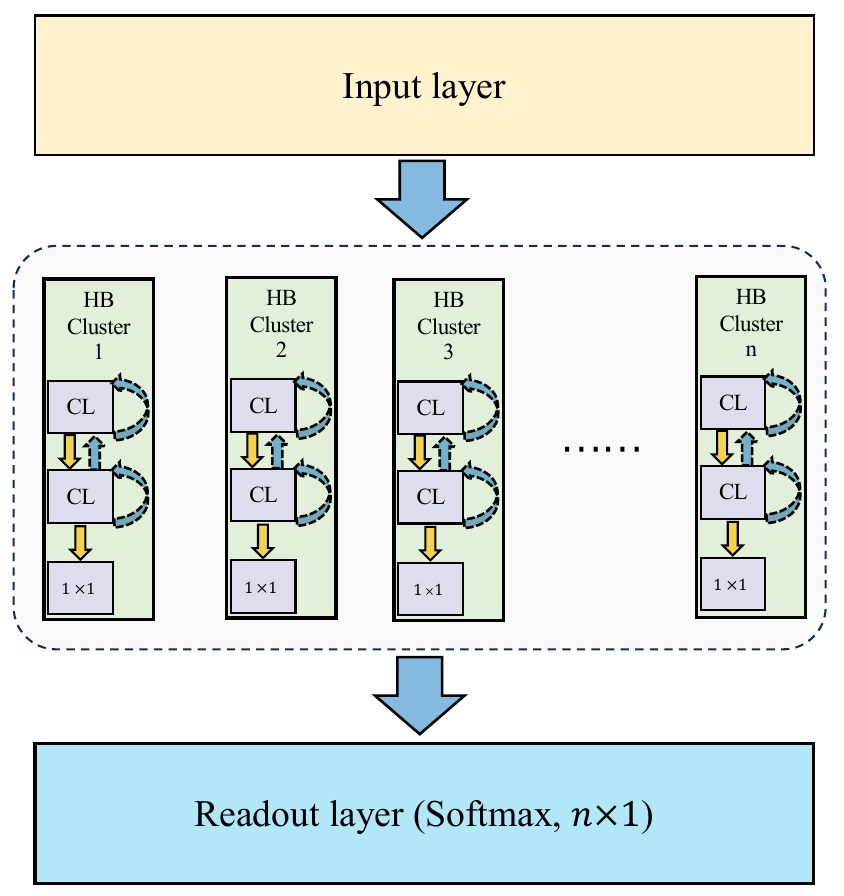}
  \caption{The architecture of HB-net consists of three primary layers: the input layer, the HB layer, and the Softmax readout layer. Notably, the HB layer is considered the core of the entire network, comprising several HB clusters. It is essential to emphasize that these HB clusters are mutually independent, with each HB cluster primarily tasked with learning the features of an individual object for recognition. In the figure, "CL" represents the "Convolutional Layer."}
  \label{HB_CNN} 
\end{figure}

HB cells are distinguished by their unique responsiveness to specific stimuli, indicating a specialized sensitivity to particular inputs. 
While there is currently no direct evidence to suggest that the visual cortex in animals mirrors the structure and mechanisms of HB cell populations found in the auditory cortex, this discovery offers valuable insights for designing networks tailored to multi-label recognition. 
To leverage this concept, distinct network modules can be dedicated to recognizing each specific label, and their individual outcomes can be amalgamated via a softmax network. 
This approach serves as the foundation for a novel framework proposed in this paper, with the overarching goal of enhancing the accuracy of multi-label recognition, even in the presence of occlusion. 
A schematic representation of this framework is depicted in Fig. \ref{HB_CNN}.

The fundamental concept underpinning this framework is the design of network modules that emulate the characteristics of HB cells. 
Each of these specialized modules is tasked with the recognition of a specific label. 
Consequently, every module within the network learns and captures the distinctive features and patterns relevant to its assigned label. 
The outputs of these individual modules are harmonized using a Softmax network, culminating in the attainment of the definitive multi-label recognition outcomes. 
The pivotal advantage of adopting this framework lies in its capacity to harness the specialization of network modules. 
Each module is meticulously trained to exhibit sensitivity and adaptability to a specific label, thereby enhancing the precision of the recognition process for the corresponding object. 
Through the integration of these module outputs, comprehensive and highly accurate multi-label recognition results can be achieved.

The network architecture comprises three primary components: a shared input layer, a shared output layer, and multiple HB cell clusters. 
Each HB cell cluster is associated with a specific label and is constituted by a neural network with a dedicated structure. 
These HB cell clusters can be constructed utilizing various network configurations, such as bottom-up (B), bottom-up with lateral feedback (BL), and bottom-up with both lateral and top-down connections (BLT) \cite{Spoerer2017}. 
In this study, the performance of these distinct network configurations on conventional occluded datasets and in the presence of noise were assessed to evaluate their effectiveness in handling complex recognition tasks.

Within these network configurations, the unidirectional feedforward connection mode serves to efficiently transmit information and systematically extract features layer by layer. 
The incorporation of recurrent connections with lateral feedback introduces valuable contextual information, thereby enhancing the network's perception and understanding of input data. 
Meanwhile, the utilization of recurrent connections with both lateral and top-down connections extends the network's memory capacity and facilitates deeper feature extraction, ultimately leading to a substantial enhancement in the network's overall performance and capability.
 
Through the comprehensive evaluation of these diverse network configurations on occluded datasets and under noisy conditions, we aim to gain deeper insights into their adaptability and robustness within intricate environments. 
This examination will furnish valuable guidance for selecting the most appropriate network configuration tailored to specific tasks, ultimately delivering effective solutions to the challenges presented by occlusion and noise in multi-label recognition tasks. 
By continually refining the network structure, we can consistently elevate the accuracy and reliability of multi-label recognition, better aligning it with the demands of practical applications and real-world scenarios.

\subsection{Different types of HB Cluster}

\begin{figure}[htbp]
  \centering
  {\includegraphics[width=0.45\textwidth]{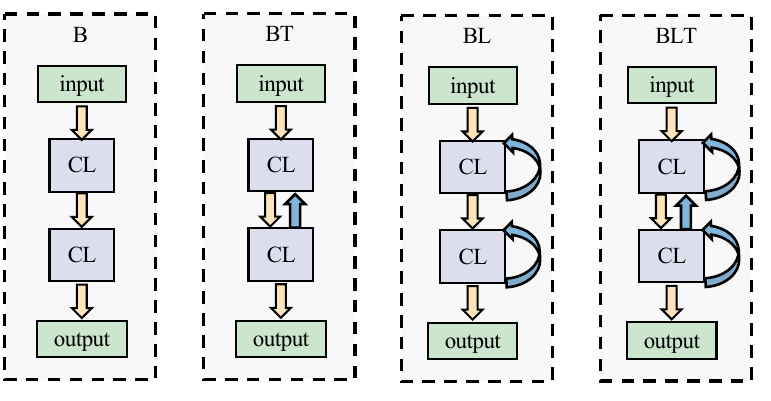}}
  \caption{Different types of HB cluster.}
  \label{Schematic}  
\end{figure}

As previously mentioned, the tested HB cell clusters are categorized into four distinct structures: B, BL, BT, and BLT, as illustrated in Fig. \ref{Schematic}. 
All the examined HB cell clusters share a common architecture, featuring two hidden recurrent convolutional layers and a single readout layer. 
The bottom-up connections are established through conventional convolutional layers with a stride of $1\times1$ \cite{Ernst2021}, while the readout layer is designed as a global average pooling layer.

To accommodate the need for different size in the top-down and lateral connections (BL, BT, BLT), the transpose convolution (also known as deconvolution) is used.
This technique enables the expansion of the top-down input to align with the dimensions of the first hidden layer. 
The connections in this layer can be conceptualized as akin to a conventional convolutional layer but with the input and output directions effectively reversed, allowing for the required adjustments in size and information flow \cite{Spoerer2017}.

In the case of the B model, a single time step is executed. 
However, for models featuring a recurrent structure (BL, BT, BLT), four time steps are processed. 
At each time point, the input is replicated. 
When evaluating accuracy, it is common practice to utilize the prediction results from the final time step, as they tend to exhibit the highest level of accuracy.

The central components of BL, BT, and BLT are the recurrent convolutional layers (RCL). 
The input to these layers is represented by ${\bf h}( \tau, m, i, j)$, which denotes the vectorized input of the central patch located at position $(i, j)$ in layer $m$, computed at time step $\tau$, across all feature maps indexed by $k$. ${\bf h}(\tau, 0, i, j)$ is defined as the input image to the network \cite{Spoerer2017}.

In the B model, the absence of recurrent connections results in a simplification of the recurrent convolutional layers (RCLs), effectively reducing them to standard convolutional layers.The pre-activation of a unit at position $(i, j)$ on feature map $k$ in layer $m$ at time step $\tau$ is defined as follows \cite{Cao2019,Spoerer2017}:
\begin{equation}
  {\bf z}_{\tau,m,i,j,k}=\left({\bf w}_{m,k}^{b}\right)^{T}{{\bf h}_{(\tau,m-1,i,j)}}+{\bf b}_{m,k} 
  \label{eqb}
\end{equation}
where $\tau = 0$ (since B runs only one time step), the convolutional kernel for the bottom connection is represented in vectorized format as ${\bf w}_{m,k}^{b}$, and the bias for feature map $k$ in layer $m$ is ${\bf b}_{m,k}$ \cite{Kietzmann2019,Spoerer2020}.

In BL, the lateral input is added to the pre-activation, which can be expressed as follows \cite{Kietzmann2019,Spoerer2020}:
\begin{equation}
  {\bf z}_{\tau,m,i,j,k}=\left({\bf w}_{m,k}^{b}\right)^{T}{\bf h}_{(\tau,m-1,i,j)}+\left({\bf w}_{m,k}^{l}\right)^{T}{\bf h}_{(\tau-1,m,i, j)}+{\bf b}_{m,k}
\end{equation}
The lateral input $\left({\bf w}_{m,k}^{l}\right)^{T}{\bf h}_{(\tau,m-1,i,j)} $ uses the same indexing as the bottom input in Eq.(\ref{eqb}), where ${\bf w}_{m,k}^{l}$ is the vectorized format of the lateral convolutional kernel. Since the lateral inputs depend on the outputs computed at time step $\tau-1$, they are undefined for the first time step ($\tau = 0$). Therefore, when $\tau = 0$, we set the recurrent inputs to zero vectors. This rule applies to all recurrent inputs, including the top-down inputs \cite{Kietzmann2019,Spoerer2020}.

In BT, we add the top-down input to the pre-activation instead of the lateral input. This yields:
\begin{equation}
  {\bf z}_{\tau,m,i,j,k}=\left({\bf w}_{m,k}^{b}\right)^{T}{\bf h}_{(\tau,m-1,i j)}+\left({\bf w}_{m,k}^{t}\right)^{T}{\bf h}_{(\tau-1,m+1,i ,j)}+{\bf b}_{m,k} 
\end{equation}
$\left({\bf w}_{m,k}^{t}\right)^{T}{\bf h}_{(\tau-1,m+1,i,j)}$ is the top-down input, where ${\bf w}_{m,k}^{t}$ denotes the vectorized format of the top-down convolutional kernel. In our model, the top-down connections can only originate from other hidden layers. Therefore, the top-down input is provided only when $m = 1$, otherwise it is set to a zero vector. The rules for top-down input also apply to the top-down input in BLT \cite{Kietzmann2019,Spoerer2020}.

Finally, we can add both the lateral and top-down inputs to the pre-activation, resulting in the layers we use in BLT \cite{Kietzmann2019,Spoerer2020}.
\begin{equation}
  \begin{array}{r}
  {\bf z}_{\tau, m, i, j, k}=\left(\mathbf{w}_{m, k}^b\right)^{T} \mathbf{h}_{(\tau, m-1, i, j)}+\left(\mathbf{w}_{m, k}^l\right)^{T} \mathbf{h}_{(\tau-1, m, i, j)} \\
  +\left(\mathbf{w}_{m, k}^t\right)^{T} \mathbf{h}_{(\tau-1, m+1, i, j)}+{\bf b}_{m, k}
  \end{array}
\end{equation}
The pre-activation ${\bf z}_{\tau, m, i, j, k}$ is processed through a layer of rectified linear units (ReLU).
\begin{equation}
  \sigma_z\left({\bf z}_{\tau, m, i, j, k}\right)=\max \left(\left\{0, {\bf z}_{\tau, m, i, j, k}\right\}\right)
  \end{equation}
\subsection{Evidence Accumulation Mechanism}

The evidence accumulation (EA) mechanism encompasses the iterative computation and accumulation of additional evidence throughout the recognition process, aimed at bolstering confidence in the target. 
This mechanism finds widespread application in the visual systems of primates \cite{Spoerer2020}. 
In the primate visual system, cognitive evidence about the target is gradually accumulated and refined through repetitive processing and analysis of incoming visual information, ultimately culminating in a final decision. 
This process of evidence accumulation can be likened to the recurrent transmission and weighted processing of neural signals in the brain, steadily accumulating and reinforcing the representation of the target. 
Such a mechanism equips both models and biological systems to effectively adapt to intricate visual environments and exhibit flexible responses to diverse images or visual stimuli.

In models based on recurrent neural networks, such as RCNN, the implementation of the evidence accumulation mechanism is accomplished through the introduction of recurrent connections. 
These connections facilitate the continuous propagation and feedback of information within the network. 
In this network architecture, cognitive evidence about the target is progressively amassed through multiple iterations of forward and backward propagation, with the ultimate goal of enhancing recognition accuracy. 
Therefore, in the aforementioned models featuring recurrent structures, the introduction of the evidence accumulation mechanism can be achieved by integrating the following method subsequent to the final convolutional layer \cite{Spoerer2020}:
\begin{equation}
  {\bf z}_{\tau, m, i, j, k} = {\bf z}_{\tau, m, i, j, k} + {\bf z}_{\tau - 1, m, i, j, k}
\end{equation} 

\begin{figure}[htbp]
  \centering
  {\includegraphics[width=0.45\textwidth]{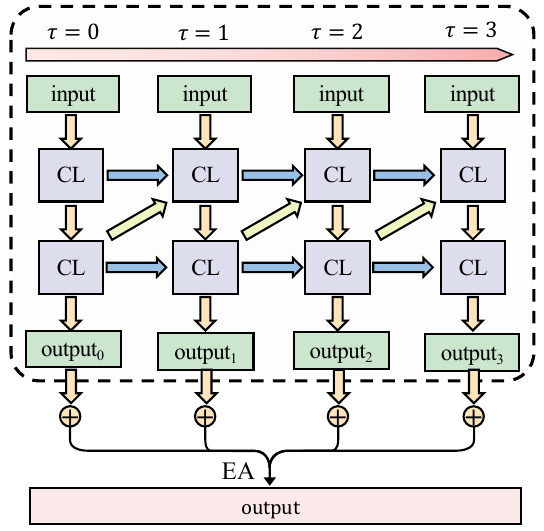}}
  \caption{A BLT model with three time steps and an EA mechanism.}
  \label{EA}  
\end{figure}

For instance, a BLT model with three time steps and an EA mechanism is depicted in Fig. \ref{EA}. The EA mechanism's function is to aggregate the outcomes produced at each time step, facilitating deeper levels of contemplation.

The BL, BT, and BLT models can seamlessly integrate the evidence accumulation mechanism. 
Upon introducing this mechanism, the model names are extended by adding the suffix "-EA" to denote the presence of the evidence accumulation mechanism. 
For instance, BL-EA signifies the BL model enhanced with the evidence accumulation mechanism.

\subsection{The read out layer and training}

Since each HB cell cluster ultimately produces a judgment value for the object, we use Softmax (Eq. \ref{softmax}) as the output layer.
\begin{equation}
  \text{Softmax}(x_i) = \frac{e^{x_i}}{\sum_{j=1}^{n}e^{x_j}}
  \label{softmax}
\end{equation}
where $x_i$ represents the $i$-th element of the input vector and $n$ represents the length of the input vector. The Softmax function calculates the exponential of each element in the input vector, normalizes the exponential results, and yields probabilities for each element. The key characteristic of the Softmax function is its ability to transform real-valued inputs into a probability distribution, ensuring that the probabilities of all categories sum up to 1. Therefore, in multi-class classification problems, the Softmax function is commonly used in the output layer to represent the predicted probabilities for each class \cite{Xia2023}.

For multi-label recognition tasks, we choose to use the binary cross-entropy loss function (BCEloss).
\begin{equation}
  \text{BCELoss}(y, \hat{y}) = -\frac{1}{N} \sum_{i=1}^{N} \left( y_i \log(\hat{y}_i) + (1 - y_i) \log(1 - \hat{y}_i) \right)
  \label{bceloss}
\end{equation}
where, $y_i$ represents the true binary label (one-hot coded) for the $i$-th sample in the dataset, which can take a value of either 0 or 1, indicating the absence or presence of a particular class or category. 
$\hat{y}_i$ represents the predicted probability for the $i$-th sample belonging to the positive class (class 1), which takes a value between 0 and 1, indicating the model's confidence in the presence of the positive class.
$N$ represents the total number of samples in the dataset.
The BCELoss function calculates the average loss over all the samples in the dataset by summing up the individual losses for each sample. 
The loss for each sample is computed using the binary cross-entropy formula.

For each sample, the binary cross-entropy loss consists of two terms: 
\begin{itemize}
  \item $y_i \log(\hat{y}_i)$: This term penalizes the model when the true label is 1 ($y_i = 1$) and the predicted probability is low ($\hat{y}_i$ is close to 0). It encourages the model to assign a high probability to the positive class.
  \item $(1 - y_i) \log(1 - \hat{y}_i)$: This term penalizes the model when the true label is 0 ($y_i = 0$) and the predicted probability is high ($\hat{y}_i$ is close to 1). It encourages the model to assign a low probability to the positive class.
\end{itemize}

The BCELoss function calculates the average of these individual losses across all the samples, yielding a single scalar value that encapsulates the overall performance of the model in the context of classification. 
The objective during the training process is to minimize this loss, thereby enhancing the model's capability to accurately categorize samples into their respective binary classes.

Throughout the training process, the Adaptive Moment Estimation (Adam) optimizer for iteration is adopted, with a weight decay set to $1e-5$ and a initial learning rate of $0.01$. 
The cosine learning rate decay is utilized to setup hyperparameters $T_{max} = 10$. 
\section[]{Experiments Design}   \label{experiment}

To illustrate the efficacy of the HB cell cluster structure, tests on HB networks featuring various types of HB cell clusters were conducted and compared with networks lacking the HB framework (ablation study). 
The network without the HB framework consists of a single cell cluster, where the output of this cell cluster signifies the probabilities of all labels rather than a single label. 
For the sake of computational simplicity, both the HB networks with HB cell cluster structures and the individual cell clusters comprise only $3$ convolutional layers. 
To distinguish between the HB cell structure and the multi-channel structure, the parameters count of which is nearly equal to the former, 
a model labeled BF (bottom-up with more features) was introduced. 
In the BF model, the number of channels in the convolutional layers is increased by a factor of $10$ or $26$, contingent on whether the datasets pertain to digits or letters. 
Furthermore, the experimental results of BF were compared with the structure of the HB network. 
Additionally, the proposed model was evaluated against existing models, including AlexNet \cite{Krizhevsky2017}, ResNet50 \cite{He2016}, and ConvNeXt-large \cite{Liu2022}.

The fundamental parameters of the models tested using the digits clutter datasets are detailed in Table \ref{tbl1}. 
The parameter format for a convolutional layer is represented as $C_i \times C_o \times k$, where $C_o$ denotes the number of output channels, $C_i$ signifies the number of input channels, and $k$ denotes the size of the convolutional kernel. 
For instance, when $k=3$, it indicates that the layer employs a $3\times3$ convolutional kernel. 
In the case of HB-net models, the format takes the shape of $n \times (C_i \times C_o \times k)$, where $n$ represents the number of object classes to be recognized. 
For digits clutter and mixed digit clutter, $n=10$, while for letters clutter, $n=26$.

\begin{table*}[htbp]
  \centering
\caption{The basic structures and parameters of the models to be tested.}\label{tbl1}
\begin{tabular*}{\linewidth}{ccccr}
    \toprule       
    \diagbox{models}{struc.} & conv1  &  conv2 &output layer &No. para \\ 
    \midrule 
    B & $1\times32\times3$  &$32\times64\times3$ & $64 \times 10  \times 3$ &24,586\\
    BF &  $1\times32\times3$  &$32\times640\times3$ & $640 \times 10  \times 3$  &242,890\\
    BT  &$1\times32\times3$  &$32\times64\times3$ & $64 \times 10  \times 3$ & 156,046  \\
    BL  &$1\times32\times3$  &$32\times64\times3$ & $64 \times 10  \times 3$  &156,046 \\
    BLT  &$1\times32\times3$  &$32\times64\times3$ & $64 \times 10  \times 3$  & 157,202\\
    BT-EA & $1\times32\times3$  &$32\times64\times3$ & $64 \times 10  \times 3$  &174,223  \\
    BL-EA & $1\times32\times3$  &$32\times64\times3$ & $64 \times 10  \times 3$  & 173,356\\
    BLT-EA & $1\times32\times3$  &$32\times64\times3$ & $64 \times 10  \times 3$  &174,512\\
    HB-B &$10\times (10\times32\times3)$  &$10 \times (32\times64\times3)$ &  $10\times(64 \times 1  \times 3)$ & 245,860\\
    HB-BL &$10\times (10\times32\times3)$  &$10 \times (32\times64\times3)$ &  $10\times(64 \times 1 \times 3)$ & 1,508,530\\
    HB-BT &$10\times (10\times32\times3)$  &$10 \times (32\times64\times3)$ &  $10\times(64 \times 1 \times 3)$ & 1,508,530\\
    HB-BLT & $10\times (1\times32\times3)$  &$10 \times (32\times64\times3)$ &  $10\times(64 \times 1  \times 3)$ & 1,520,090 \\
    HB-BL-EA & $10\times (1\times32\times3)$  &$10 \times (32\times64\times3)$ &  $10\times(64 \times 1  \times 3)$ &1,525,840 \\
    HB-BT-EA & $10\times (1\times32\times3)$  &$10 \times (32\times64\times3)$ &  $10\times(64 \times 1  \times 3)$ & 1,525,840 \\
    HB-BLT-EA & $10\times (1\times32\times3)$  &$10 \times (32\times64\times3)$ &  $10\times(64 \times 1  \times 3)$ &1,537,400 \\
    ResNet50  & - & - &-&23,522,250\\
    AlexNet  & - & - &-&57,029,322\\
    ConvNeXt-large & - & - &-&196,212,490\\
    \bottomrule   
    \end{tabular*}
\end{table*}

Finally, the performance tests on different models under various levels of Gaussian noise and different signal-to-noise ratios of salt-and-pepper noise were conducted to evaluate their robustness.

These tests observed and contrasted the performance of different models in noisy environments. 
Analyzing how the models' performance changes under diverse noise conditions provides insights into their resilience to noise and their effectiveness across varying signal-to-noise ratios.

The outcomes of these tests offer crucial insights into the models' robustness and their applicability in real-world scenarios. 
By assessing their performance in noisy environments, valuable information obtained can inform enhancements in the design and training strategies, ultimately leading to improved robustness and accuracy in practical applications.

\section[]{Results}  \label{results}

Based on the experiments conducted, the accuracy of each model for the scenarios of $5$-digit clutter, $5$-letter clutter, and mixed $5$-digit clutter are recorded in Table \ref{tbl2}, \ref{tbl3}, and \ref{tbl4}. 

\begin{table*}[h]
  \centering
  \addtolength{\tabcolsep}{-3pt}
\caption{The models results' under 5-digit clutter.}\label{tbl2}
\begin{tabular*}{\linewidth}{cc|ccc|ccc}
  \toprule  
   \multirow{2}*{Model}& \multirow{2}*{Original Data}&\multicolumn{3}{c|}{Gaussian} &\multicolumn{3}{c}{salt-and-pepper}  \\ 
  \multicolumn{2}{c|}{} & $\sigma = 0.5$& $\sigma = 3$& $\sigma = 5$& $SNR = 0.9$& $SNR = 0.5$& $SNR =0.1$\\
  \midrule 
  AlexNet & 0.8154 & 0.7727 & 0.5047 & 0.5027 & 0.7442& 0.5661 & 0.5048\\
  ResNet50 & 0.8504 & 0.8163 & 0.5362 & 0.5072 & 0.7654& 0.5872 & \textbf{0.5049} \\
  ConvNeXt-large &\textbf{0.8748} &\textbf{0.8363} &\textbf{0.5566}&\textbf{0.5178}&\textbf{0.8602}&\textbf{0.7844}&0.5029\\
  B & 0.6852 & 0.6407 & 0.5123 & 0.5045 & 0.6248 & 0.5025 &0.5031\\ 
  BF & 0.6681 & 0.6441 & 0.5155 & 0.5030 &0.6284 &0.5041 & 0.5025\\
  BT & 0.6843 & 0.6514 & 0.5157 & 0.5040 & 0.6264 &0.5040 &0.5034 \\
  BL & 0.7026 & 0.6703 & 0.5167 & 0.5042 & 0.6312&0.5040&0.5047\\
  BLT & 0.7023 & 0.6681 & 0.5144 & 0.5035 & 0.6876 &0.5042 &0.5046\\
  BT-EA & 0.7775 & 0.6792 & 0.5158 & 0.5043 & 0.6434 &0.5064 &0.5027\\
  BL-EA & 0.7837 & 0.7058 & 0.5166 & 0.5033 & 0.6617&0.5033&0.5033\\
  BLT-EA & 0.7718 & 0.7175 & 0.5128 & 0.5036 & 0.6675 &0.5033&0.5045\\
  HB-B & 0.7966 & 0.7292 & 0.5177 & 0.5055 & 0.6950 & 0.5052 & \textbf{0.5049}\\
  HB-BL & 0.8439 & 0.7873 & 0.5143 & 0.5046 & 0.7547 & 0.5039 & 0.5034\\
  HB-BT & 0.8328  & 0.7735 & 0.5170 & 0.5061 & 0.7418 &0.5036&0.5035\\
  HB-BLT & 0.8474  & 0.7904 & 0.5177 & 0.5050 & 0.7643 &0.5044& 0.5034\\
  HB-BL-EA & 0.8702  & 0.8125 & 0.5392 & 0.5142 & 0.8211 &0.5284&0.5029\\
  HB-BT-EA & 0.8543  & 0.8077 & 0.5209 &0.5117  & 0.7837 &0.5243&0.5024\\
  HB-BLT-EA & 0.8634  & 0.8126 & 0.5387& 0.5102 & 0.8232 & 0.5294 & 0.5031\\
  \bottomrule  
  \end{tabular*}
\end{table*}

\begin{table*}[htbp]
  \centering
  \addtolength{\tabcolsep}{-3pt}
\caption{The models results' under 5-letter clutter.}\label{tbl3}
\begin{tabular*}{\linewidth}{cc|ccc|ccc}
  \toprule    
     \multirow{2}*{Model}& \multirow{2}*{Original Data}&\multicolumn{3}{c|}{Gaussian} &\multicolumn{3}{c}{pepper salt}  \\ 
    \multicolumn{2}{c|}{} & $\sigma = 0.5$& $\sigma = 3$& $\sigma = 5$& $SNR = 0.9$& $SNR = 0.5$& $SNR =0.1$\\
    \midrule 
    AlexNet & 0.8932 & 0.8742 & 0.6906 & 0.6908 & 0.8589 & 0.8202& 0.6914\\
    ResNet50 & 0.9008 & 0.8855 & 0.7139 & 0.6952 & 0.8843 & 0.8235 & 0.6954\\
    ConvNeXt-large & 0.8435 &0.8178 &0.6994&0.6965&0.7648&0.7059&0.6925\\
    B & 0.8156 & 0.7872 & 0.6907 & 0.6909 & 0.7838 & 0.6911 & 0.6904\\ 
    BF & 0.8606 & 0.8285 & 0.7130 & 0.6990 & 0.7724 & 0.6913 & 0.6902\\
    BT & 0.8230 & 0.8033 & 0.7031 & 0.6908 & 0.7954 & 0.6969 & 0.6911\\
    BL & \textbf{0.9254} & 0.8832 & \textbf{0.7219} & \textbf{0.6992} & 0.8133 & 0.7989 & 0.6912\\
    BLT & 0.9218 & 0.8838 & 0.7190 & 0.6984 & 0.8267 & 0.7899 & 0.6915\\
    BT-EA & 0.8950 & 0.8591 & 0.7149 & 0.6978 & 0.8155 & 0.7856 & 0.6903\\
    BL-EA & 0.9218 & 0.8843& 0.7165 & 0.6983 & 0.8234 & 0.7924 & 0.6908\\
    BLT-EA & 0.9194 & 0.8858 & 0.7162 & 0.6975 & 0.8298 & 0.7967 & 0.6904\\
    HB-B & 0.8792 & 0.8401 & 0.7098 & 0.6906 & 0.8233 & 0.8044 & 0.6911\\
    HB-BL & 0.9201 & \textbf{0.8866} & 0.7032 & 0.6986 & 0.8817 & 0.8109 & 0.6954 \\
    HB-BT & 0.9053  & 0.8719 & 0.7101 & 0.6985 & 0.8749 & 0.8134 & 0.6978\\
    HB-BLT & 0.9151  & 0.8820 & 0.7030 & 0.6989 & 0.8922 & 0.8156 & 0.6952\\
    HB-BL-EA & 0.9184  & 0.8845 & 0.7134 & 0.6970& 0.8942 & \textbf{0.8313} & \textbf{0.6984}\\
    HB-BT-EA & 0.9083  & 0.8700& 0.7126 & 0.6987  & 0.8644 & 0.8105 & 0.6932\\
    HB-BLT-EA & 0.9155  & 0.8800 & 0.7128& 0.6972  & \textbf{0.9012}& 0.8206 & 0.6945 \\
    \bottomrule    
  \end{tabular*}
\end{table*}

\begin{table*}[htbp]
  \centering
  \addtolength{\tabcolsep}{-3pt}
\caption{The models results' under mixed 5-digit clutter.}\label{tbl4}
\begin{tabular*}{\linewidth}{cc|ccc|ccc}
  \toprule    
     \multirow{2}*{Model}& \multirow{2}*{Original Data}&\multicolumn{3}{c|}{Gaussian} &\multicolumn{3}{c}{salt-and-pepper}  \\ 
    \multicolumn{2}{c|}{} & $\sigma = 0.5$& $\sigma = 3$& $\sigma = 5$& $SNR = 0.9$& $SNR = 0.5$& $SNR =0.1$\\
    \midrule 
    AlexNet & 0.8108 & 0.7276 & 0.6677 & 0.6201 & 0.7377 & 0.5468 & 0.5035 \\
    ResNet50 & 0.8604 & \textbf{0.8238} & 0.6629 & 0.6234 & 0.7689 & \textbf{0.6043} & 0.5048\\
    ConvNeXt-large &0.8709&0.8210 &0.6385&0.6278&0.7600&0.5843&0.5029\\
    B & 0.7836 & 0.7451 & 0.6558 & 0.6236 & 0.6296 &0.5264 &0.5039\\ 
    BF &0.7925&0.7587&0.6342& 0.6119& 0.6384& 0.5099& 0.5045\\
    BT & 0.7889 & 0.7647 & 0.6447 & 0.6240 & 0.6583&0.5048 &0.5032\\
    BL &0.8071& 0.7750& 0.6780&0.6285& 0.6648&0.5110& 0.5053\\
    BLT &0.8036& 0.7746& 0.6835& 0.6200& 0.6604& 0.5200& 0.5008\\
    BT-EA &0.8047& 0.7835& 0.6739& 0.6235& 0.6681& 0.5107& 0.5005\\
    BL-EA &0.8289& 0.7904& 0.6972& 0.6223& 0.6790& 0.5118& 0.5029\\
    BLT-EA &0.8275&0.8054& 0.6888&0.6395& 0.6811&0.5292& 0.5039\\
    HB-B &0.8244& 0.7903& 0.6820&0.6245&0.6972& 0.5294& 0.5045\\
    HB-BL  &0.8509& 0.8099&0.6944& 0.6299& 0.7547&0.5208& 0.5021\\
    HB-BT &0.8373& 0.8044& 0.6769& 0.6218& 0.7457& 0.5054& 0.5008\\
    HB-BLT &0.8605& 0.8145&0.6987&0.6290& 0.7570& 0.5279& 0.5029\\
    HB-BL-EA &\textbf{0.8789}& 0.8120& 0.7125&0.6321& 0.8010&0.5192& \textbf{0.5059}\\
    HB-BT-EA &0.8481& 0.8068& 0.7061& 0.6246&0.7730& 0.5187&0.5047\\
    HB-BLT-EA &0.8634& 0.8221&\textbf{0.7227}& \textbf{0.6396}& \textbf{0.8247}&0.5267& 0.5039\\
    \bottomrule 
  \end{tabular*}
\end{table*}

Here is some initial observations made from these tables:

\begin{itemize}
  \item The proposed models can be trained on various datasets and noise levels. 
  Notably, the highest accuracy is attained on the $5$-letter clutter dataset, followed by the mixed $5$-digit clutter dataset, and lastly, the $5$-digit clutter dataset. 
  As expected, with the escalation of noise levels, the recognition accuracy of each model progressively diminishes. 
  The degree of accuracy reduction serves as an indicator of the model's robustness to noise.
  \item The average accuracy of the HB-BLT-EA, HB-BL-EA, ConvNeXt-large, and ResNet50 models is notably high, with minimal variation between them.
  \item In contrast to models lacking the HB framework, models that integrate the HB framework consistently exhibit improved accuracy across all three datasets. 
  A preliminary comparison between models BF and HB, which have similar parameter counts, suggests that the presence of the HB framework contributes to an increase in model accuracy, rather than a growth in the number of model parameters. 
  Moreover, when comparing models B and BF, it is apparent that model BF achieves slightly higher accuracy, though the improvement is not substantial. 
  This implies that merely augmenting the model's channel count to expand its width and increase the parameter count does not significantly enhance the model's accuracy in the context of multi-label recognition. 
  This observation further supports the effectiveness of the HB framework.
 \item In comparison to models without an EA mechanism, models equipped with an EA mechanism have also demonstrated improved accuracy across all three datasets.
  \item Among the three datasets, the top-performing HB-BLT-EA and HB-BL-EA models achieve accuracy levels comparable to ResNet50 and ConvNeXt-large, and in some cases, even surpass the training results of ResNet. 
  Notably, the parameter count of the HB-BL-EA model is only about $1/30$ and $1/100$ that of ResNet50 and ConvNeXt-large respectively, consisting of only two convolutional layers. 
  This remarkable performance with significantly fewer parameters underscores the efficiency and effectiveness of the proposed HB framework and EA mechanism.
\end{itemize}

\section{Discussion}   \label{discussion}

To demonstrate the effectiveness of the proposed framework and the robustness of the models against two types of noise, a comparative analysis was conducted using the Wilcoxon signed-rank test on the aforementioned results. 
The Wilcoxon signed-rank test is a non-parametric hypothesis test used to compare differences between two related samples. 
It is suitable for situations where the sample data does not follow a normal distribution assumption or when other parametric tests cannot be applied.

The Wilcoxon signed-rank test is based on the ranking of the absolute differences between paired samples. 
It calculates the sum of ranks for positive and negative differences separately and evaluates the difference between the two samples by comparing the difference between these sums of ranks. 
The null hypothesis of the Wilcoxon signed-rank test assumes that there is no significant difference between the paired samples, while the alternative hypothesis suggests the presence of a difference. 
Statistical conclusions about the significance of the difference are drawn by comparing the calculated sum of ranks with the expected values.

Given the limited number of test results in this study, which makes it challenging to determine their distribution, the Wilcoxon test was employed for comparing the experimental results within each group. 
The goal was to obtain statistically significant comparisons. 
This section initially highlights the effectiveness of the HB-net framework and the EA mechanism. Following that, it compares the robustness of various models.

\subsection{effectiveness}

The effectiveness of the HB-net framework and EA mechanism were evaluated using the Wilcoxon test, which aimed at a thorough comparative analysis.

\subsubsection[]{HB framework}

The effectiveness of the HB framework was confirmed through a Wilcoxon signed-rank test, comparing the results of $147$ models generated by the HB framework with the corresponding models lacking the HB framework (excluding the BF model). 
The test results revealed a significant difference between the model results with the HB framework and those without it ($p=0.000$). 
The models incorporating the HB framework exhibited a $2.98\%$ improvement in accuracy compared to their counterparts lacking the framework ($p=0.0499$).

A comparison was made between the BF model and the HB-B model, which have similar parameter sizes. 
Utilizing a Wilcoxon signed-rank test on the results from $21$ tests conducted across various datasets and noise levels, it was evident that there is a significant difference between the BF model and the HB-B model ($p=0.000$). 
The HB-B model showed a notable improvement in accuracy, achieving a $2.59\%$ increase compared to the BF model ($p=0.0479$).

The comparisons between models with the HB framework and their corresponding models without the HB framework, as detailed in Table \ref{compare_hb}, consistently show that the HB models exhibit a significant ($p<0.05$) improvement in accuracy compared to their non-HB counterparts. 
This pattern is also evident in Table \ref{compare_EA} and Table \ref{compare_alex}, where a p-value below $0.05$ indicates a significant accuracy improvement.

\begin{table*}[htbp]
  \addtolength{\tabcolsep}{-3pt}
\caption{The Wilcoxon test results of HB and non-HB models.}\label{compare_hb}
\begin{tabular*}{\linewidth}{cccccccc}
  \toprule      
   Non-HB model & HB model &No. of results& $Stat.$ & $p$ & Increment(\%) & $Stat._{in}$ & $p_{in}$ \\ 
  \midrule 
  Non-HB & HB & 147 & 810 & 2.80e-19& 2.98 & 4588 & 0.0499\\
  BF & HB-B &21& 11 & 0.0002 & 2.59 & 67 & 0.0479\\
  B & HB-B &21 & 0 & 4.77e-7 & 2.96 & 67 & 0.0479 \\
  BL & HB-BL &21 & 42 & 0.0045 & 0.42 & 67 & 0.0479 \\
  BT & HB-BT &21 & 15 & 6.53e-5 & 3.10& 67 & 0.0479 \\
  BLT & HB-BLT &21 &24.5 & 0.0004 & 0.82 & 67 & 0.0479 \\
  BT-EA & HB-BT-EA &21 & 5 & 4.77e-6 & 1.52 & 67 & 0.0479 \\
  BL-EA & HB-BL-EA &21 & 16 & 8.06e-5 & 1.65 & 67 & 0.0479 \\
  BLT-EA & HB-BLT-EA &21 & 28 & 0.0020 & 1.44 & 67 & 0.0479 \\
  \bottomrule     
  \end{tabular*}
\end{table*}

\subsubsection[]{EA structure}

The effectiveness of the EA structure was also evaluated. 
Similar to the HB framework, a series of tests were conducted to assess the performance of models with the EA structure. 
The results are presented in Table \ref{compare_EA}, where the EA models are compared with their corresponding models without the EA structures. 
This analysis aims to demonstrate the effectiveness of the EA structure in improving model performance.

\begin{table*}[htbp]
  \addtolength{\tabcolsep}{-3pt}
  \caption{The Wilcoxon test results of EA and non-EA models.}\label{compare_EA}
  \begin{tabular*}{\linewidth}{cccccccc}
    \toprule      
     Non-EA model & EA model &No. of results& $Stat.$ & $p$ & Increment(\%) & $Stat._{in}$ & $p_{in}$ \\ 
 
    \midrule 
    Non-EA & EA & 126 & 1763.5 & 4.24e-8& 0.86 & 3323 & 0.0495\\
    BT & BT-EA &21& 19 & 0.0001 & 1.25 & 67 & 0.0479\\
    BL & BL-EA &21 & 80 & 0.1145 & -& - & - \\
    BLT & BLT-EA &21 & 52.5 & 0.0132 & 0.15& 67 & 0.0479 \\
    HB-BT & HB-BT-EA &21 & 37 & 0.0024 & 0.33 & 67 & 0.0479 \\
    HB-BL & HB-BL-EA &21 & 21.5 & 0.0002 & 0.89 & 67 & 0.0479 \\
    HB-BLT & HB-BLT-EA &21 & 31 & 0.0011 & 0.47 & 67 & 0.0479 \\
    \bottomrule   
    \end{tabular*}
  \end{table*}

The results from Table \ref{compare_EA} unequivocally reveal that, overall, models integrated with the EA mechanism exhibit a substantial increase in recognition accuracy compared to models lacking the EA structure, across all three datasets ($p=0.000$). 
Models featuring the EA mechanism demonstrate a notable $0.86\%$ improvement in accuracy when contrasted with models without the EA mechanism ($p=0.0495$). 
With the exception of BL-EA, which did not exhibit significant accuracy enhancements compared to BL ($p=0.1145$), all other models incorporating the EA mechanism displayed a statistically significant accuracy enhancement compared to their counterparts without the EA structure ($p<0.05$).

\subsubsection{Overall}

Based on the results obtained from various models across different datasets, it is evident that models incorporating the HB framework and EA mechanism consistently outperform other models in terms of recognition accuracy. 
In order to provide a comparison with commonly used models, tests on the AlexNet, ResNet50, and ConvNeXt-large models under varying noise levels across the three datasets were conducted. 
Table \ref{compare_alex} highlights the improvements in accuracy achieved by the HB-BLT-EA and HB-BL-EA models relative to the AlexNet, ResNet50, ConvNeXt-large, B, and BF models.

\begin{table*}[htbp]
  \addtolength{\tabcolsep}{-3pt}
  \caption{The Wilcoxon test results of some important models.}\label{compare_alex}
  \begin{tabular*}{\linewidth}{cccccccc}
    \toprule   
     model(s)1 & model(s)2 &No. of results& $Stat.$ & $p$ & Increment(\%) & $Stat._{in}$ & $p_{in}$ \\ 
    \midrule  
    B & BF &21 & 88.5 & 0.1777 & - & - & - \\
    B & HB-BLT-EA &21 & 0 & 6.59e-5 & 6.17 & 67 & 0.0479 \\
    BF & HB-BLT-EA &21& 7 & 9.06e-6 & 4.67 & 67 & 0.0479\\
    AlexNet & HB-BLT-EA &21& 25 & 0.0004 & 1.62 & 67 & 0.0479\\
    ResNet50 & HB-BLT-EA &21& 85.5 & 0.1519 & - & - & -\\
    ConvNeXt-large & HB-BLT-EA &21& 101 & 0.4406 & - & - & -\\
    HB-BL-EA & HB-BLT-EA &21& 110 & 0.4324 & - & - & -\\
    B & HB-BL-EA &21 & 5 & 4.77e-6 &5.72 & 67 & 0.0479 \\
    BF & HB-BL-EA &21& 4 & 3.34e-6 & 4.60 & 67 & 0.0479\\
    AlexNet & HB-BL-EA &21& 26 & 0.0005 & 1.94 & 67 & 0.0479\\
    ResNet50 & HB-BL-EA &21& 70 & 0.0597 & - & - & -\\
    ConvNeXt-large & HB-BL-EA &21& 101 & 0.4406& - & -& -\\
    \bottomrule    
    \end{tabular*}
  \end{table*}

The results presented in Table \ref{compare_alex} indicate that the HB-BL-EA and HB-BLT-EA models achieve a significant improvement ($p<0.05$) in recognition accuracy when compared to B, BF, and AlexNet. 
Notably, there is no statistically significant difference in accuracy between models B and BF across the three datasets ($p=0.1777>0.05$). 
This suggests that the enhanced accuracy of the HB-B model is primarily attributed to the presence of the HB framework, rather than a mere increase in model width.

Furthermore, it is evident that HB-BL-EA, HB-BLT-EA, ConvNeXt-large, and ResNet50 exhibit no significant differences in accuracy among the four models. 
This suggests that models equipped with both the HB framework, EA mechanism, and lateral recurrent connections can achieve a performance level similar to that of ResNet50. 
However, it is important to note that HB-BL-EA and HB-BLT-EA have approximately $1/30$ the parameter count of ResNet50 and consist of only three layers, making them more efficient in terms of model complexity.

\subsection{Robustness}
To assess the robustness of different models to noise, the comparison of the extent of accuracy degradation under varying noise levels provides insights into their noise tolerance. 
For example, the robustness of the HB framework was evaluated in comparison to models without the HB framework under Gaussian noise. 
The accuracy data of the HB framework on the original dataset and at different noise levels were collected, denoted as $A_{\text{HB}}$. 
Then, the difference in accuracy between adjacent noise levels for the HB framework can be calculated by:
\begin{eqnarray}
  \Delta_0 A_{\text{HB},(\sigma = 0, \sigma = 0.5)}& = &A_{\text{HB},\sigma = 0} - A_{\text{HB}, \sigma = 0.5} \nonumber\\
  \Delta_1 A_{\text{HB},(\sigma = 0.5, \sigma = 3)} &= &A_{\text{HB},\sigma = 0.5} - A_{\text{HB}, \sigma = 3}\\
  \Delta_2 A_{\text{HB},(\sigma = 3, \sigma = 3)} &= &A_{\text{HB},\sigma = 3} - A_{\text{HB}, \sigma = 5}\nonumber
\end{eqnarray}

Likewise, the accuracy data of the non-HB framework on the original dataset and different levels of noise were also calculated denoted as $A_{\overline{\text{HB}}}$, and then the difference in accuracy between adjacent noise levels for the non-HB framework can be obtained.
\begin{eqnarray}
  \Delta_0 A_{\overline{\text{HB}},(\sigma = 0, \sigma = 0.5)}& = &A_{\overline{\text{HB}},\sigma = 0} - A_{\overline{\text{HB}}, \sigma = 0.5} \nonumber\\
  \Delta_1 A_{\overline{\text{HB}},(\sigma = 0.5, \sigma = 3)} &= &A_{\overline{\text{HB}},\sigma = 0.5} - A_{\overline{\text{HB}}, \sigma = 3}\\
  \Delta_2 A_{\overline{\text{HB}},(\sigma = 3, \sigma = 3)} &= &A_{\overline{\text{HB}},\sigma = 3} - A_{\overline{\text{HB}}, \sigma = 5}\nonumber
\end{eqnarray}

Compare the robustness of the HB framework and the non-HB framework to Gaussian noise can be carried out by the Wilcoxon signed-rank test on $[\Delta_0 A_{\text{HB},(\sigma = 0, \sigma = 0.5)}, \Delta_1 A_{\text{HB},(\sigma = 0.5, \sigma = 3)}, $ $\Delta_2 A_{\text{HB},(\sigma = 3, \sigma = 5)}]$ 
and $[\Delta_0 A_{\overline{\text{HB}},(\sigma = 0, \sigma = 0.5)},$ $\Delta_1 A_{\overline{\text{HB}},(\sigma = 0.5, \sigma = 3)},\Delta_2 A_{\overline{\text{HB}},(\sigma = 3, \sigma = 5)}]$. 

To perform the robustness comparison among different models under Gaussian and salt-and-pepper noise, differential vectors of accuracy for each model and category were calculated. 
These differential vectors represent the change in accuracy under different noise levels. 
After obtaining the accuracy differential vectors for each model and category, paired Wilcoxon tests were used to compare the robustness among different categories of models (HB-nets vs. their non-HB counterparts, models with vs. without the EA framework) and between different models (e.g., ResNet50 vs. HB-BL-EA, etc.). 
The significant robustness comparison results will be presented in Tables \ref{compare_hb_r}, \ref{compare_EA_r}, and \ref{compare_alex_r}, with p-values indicating whether there is a significant difference in robustness.

\begin{table*}[htbp]
  \caption{The Wilcoxon test results for the rubustness of non-HB and HB models.}\label{compare_hb_r}
  \centering
  \begin{tabular*}{0.7\linewidth}{ccccc}
    \toprule     
     Non-HB model & HB model &No. of results& $Stat.$ & $p$  \\ 
    \midrule  
    Non-HB  & HB & 126 & 2152.5 & 3.41e-6\\
    B & HB-B &18 & 33 & 0.0104 \\
    BF & HB-B &18 & 50 & \textbf{0.0648} \\
    BL & HB-BL &18 & 39 & 0.0216 \\
    BT & HB-BT &18 & 30 & 0.0069  \\
    BLT & HB-BLT &18 & 36 & 0.0152 \\
    BT-EA & HB-BT-EA &18 & 63 & \textbf{0.1733}  \\
    BL-EA & HB-BL-EA &18 & 54 & \textbf{0.0907} \\
    BLT-EA & HB-BLT-EA &18 & 64 & \textbf{0.1846} \\
    \bottomrule     
    \end{tabular*}
  \end{table*}
  
  \begin{table*}[htbp]
    \caption{The Wilcoxon test results for the rubustness of Non-EA and EA models.}\label{compare_EA_r}
    \centering
    \begin{tabular*}{0.7\linewidth}{ccccc}
      \toprule    
       Non-EA model & EA model &No. of results& $Stat.$ & $p$ \\ 
      \midrule  
      Non-EA & EA & 108 & 1938.5 & 0.0010\\
      BT & BT-EA &18& 27 & 0.0045 \\
      BL & BL-EA &18 & 35 & 0.0134  \\
      BLT & BLT-EA &18 & 61 & \textbf{0.1519}  \\
      HB-BT & HB-BT-EA &18 &64 & \textbf{0.1846}  \\ 
      HB-BL & HB-BL-EA &18 & 65 & \textbf{0.1964} \\
      HB-BLT & HB-BLT-EA &18 & 80 & \textbf{0.4159} \\
      \bottomrule     
      \end{tabular*}
    \end{table*}

    \begin{table*}[htbp]
      \caption{The Wilcoxon test results for the rubustness of some important models.}\label{compare_alex_r}
      \centering
      \begin{tabular*}{0.7\linewidth}{ccccc}
        \toprule     
         model(s)1 & model(s)2 &No. of results& $Stat.$ & $p$ \\ 
        \midrule 
        B & BF &18 & 74 & \textbf{0.6397} \\
        B & HB-BLT-EA &18 & 37 & 0.0171 \\
        BF & HB-BLT-EA &18& 38 & 0.0192 \\
        AlexNet & HB-BLT-EA &18& 69.5 & \textbf{0.4951}\\
        ResNet50 & HB-BLT-EA &18& 79 & \textbf{0.7987} \\
        ConvNeXt-large & HB-BLT-EA &18& 73 & \textbf{0.6095}\\
        HB-BL-EA & HB-BLT-EA &18& 75 & \textbf{0.6705 }\\
        B & HB-BL-EA &18 & 35 & 0.0134 \\
        BF & HB-BL-EA &18& 41 & 0.0269 \\
        AlexNet & HB-BL-EA &18& 56 & \textbf{0.2121} \\
        ResNet50 & HB-BL-EA &18& 79 & \textbf{0.7987} \\
        ConvNeXt-large & HB-BL-EA &18& 66 & \textbf{0.4171}\\
        \bottomrule   
        \end{tabular*}
      \end{table*}
  \begin{figure*}[htbp]
    \centering
    {\includegraphics[width=0.9\textwidth]{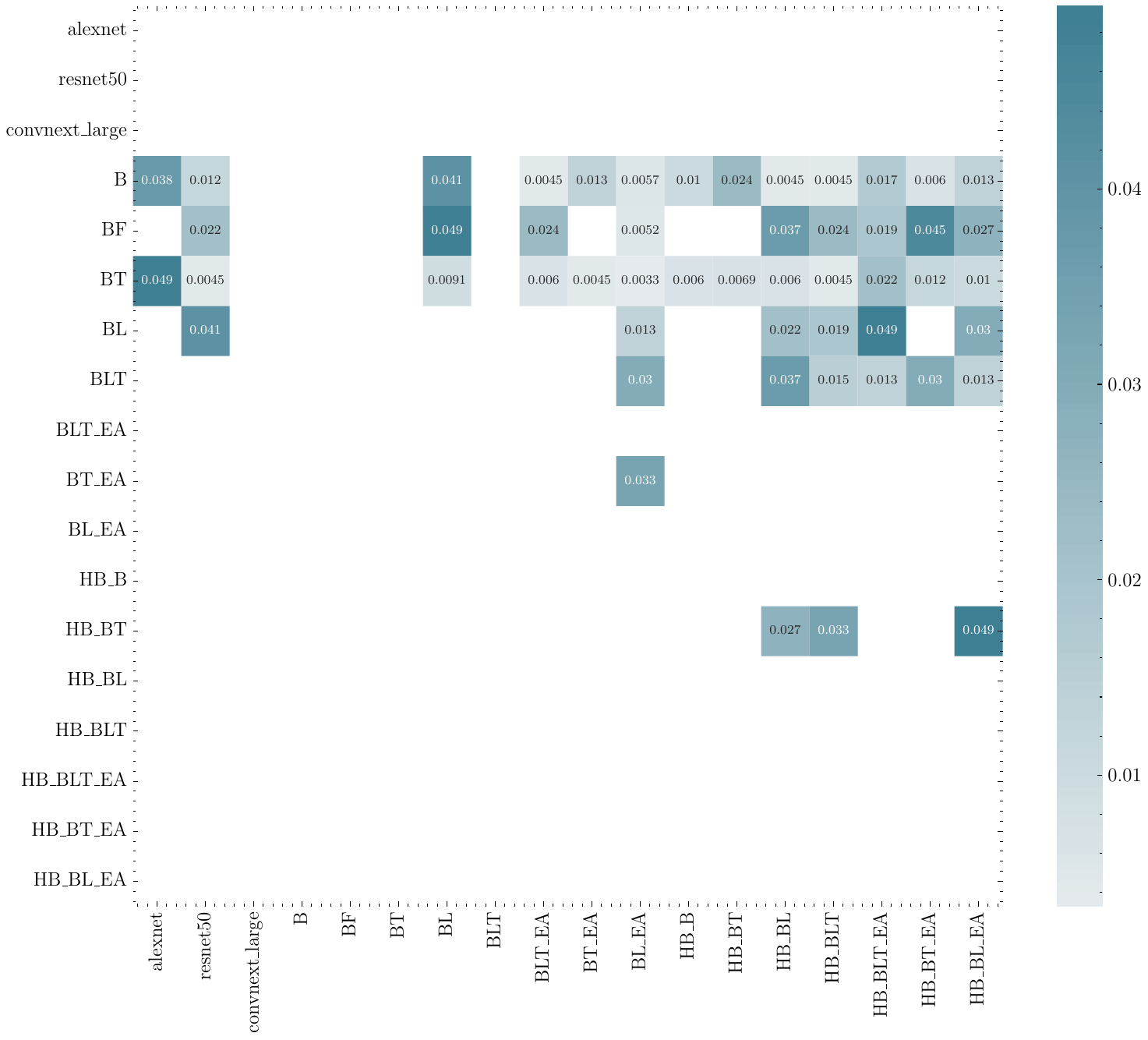}}
    \caption{Statistical comparison between our models and other approaches (p-value matrix).}
    \label{robustness}  
  \end{figure*}

In these tables, if a row's p-value is not in bold (i.e., $p < 0.05$), it suggests that there is a significant difference in robustness between the models corresponding to non-HB/EA models and HB/EA models under the two types of noise. 
Conversely, if the p-value is in bold, it indicates that there is no significant difference in robustness between the models under both types of noise.

Additionally, the robustness comparison results among different models are depicted in Fig. \ref{robustness}, where only p-values indicating significant differences in robustness will be displayed. If a cell contains a p-value, it signifies that the model corresponding to the row is more robust than the model corresponding to the column, and this difference is statistically significant ($p < 0.05$).

From the robustness comparison results presented in Tables \ref{compare_hb_r}, \ref{compare_EA_r}, and \ref{compare_alex_r}, and the insights from Fig. \ref{robustness}, several key observations can be made:
\begin{itemize}
  \item Models without the HB framework (B and BF) tend to exhibit better robustness compared to HB-net models (HB-B, HB-BL, etc.). 
  This suggests that the HB framework may introduce some instability into the models that affects their robustness. 
  In particular, models with the EA mechanism and the HB framework are less robust compared to those without the HB framework.
  \item Models without the EA mechanism generally display better robustness compared to models with the EA mechanism. 
  This trend holds across various HB-net models. 
  However, it's notable that there is no significant difference in robustness between HB-BLT and HB-BLT-EA.
  \item A trade-off between a model's robustness and accuracy can be observed. 
  Higher accuracy models tend to have lower robustness. 
  When noise levels are extremely high, most models struggle to extract meaningful information from the data, leading to accuracy levels close to random guessing and increased variations in accuracy differences.
\end{itemize}
 
In summary, the results suggest that while the HB framework and the EA mechanism may enhance accuracy, they can compromise robustness, particularly when facing high levels of noise. 
The choice between these model structures should be guided by the specific needs of the application and the trade-off between accuracy and robustness. 
Models without the HB framework and the EA mechanism can offer better robustness, making them suitable for tasks where robustness is a priority. 
Conversely, models with the HB framework and EA mechanism may be more appropriate for tasks where high accuracy is essential, and robustness against noise is a lesser concern.

\section{Conclusions}   \label{conclusion}

This work introduces the HB-net framework, which leverages Holistic Bursting (HB) cell clusters to address the challenge of multi-object occlusion in images. 
It explores various HB cell cluster structures, including bottom-up (B), bottom-up with lateral feedback (BL), and bottom-up with lateral and top-down connections (BLT). 
Additionally, an evidence accumulation mechanism is incorporated. 
Evaluation on datasets containing digits and letters demonstrates that models within the HB framework exhibit a significant $2.98\%$ improvement in recognition accuracy compared to conventional convolutional networks ($p=0.0499$). 
Furthermore, models with the evidence accumulation mechanism show a $0.86\%$ increase in recognition accuracy compared to their counterparts without this mechanism ($p=0.0495$). 
However, under high noise levels, conventional convolutional networks demonstrate higher robustness compared to HB-net models.

Potential future research directions are as follows:
\begin{itemize}
  \item While the study has explored different recurrent connections (BL, BLT, B, BT) and evidence accumulation mechanisms (EA) within the HB framework, there is room to investigate the integration of various basic convolutional blocks as HB cell clusters. 
  This includes considering structures like residual blocks similar to those in ResNet or inception blocks like those in GoogLeNet.
  \item The study discusses using the same HB cell cluster structure within each HB-net. 
  Future research could explore the use of heterogeneous HB cell clusters tailored to specific problem characteristics to enhance recognition performance.
  \item The integration of attention mechanisms into HB-nets or EA process presents an intriguing avenue for improving recognition performance, given the promising results achieved by attention mechanisms in recent years.
  \item While simulated datasets were used to validate the advantages of the HB framework in scenarios involving occlusion, applying HB-net to real datasets presents an important future research direction. 
  Real datasets differ in factors like resolution, background, and noise, and assessing HB-net's performance on these datasets is crucial.
  \item Researching how to efficiently recognize a large number of different object categories within the same field of view is a meaningful challenge. 
  It involves addressing model efficiency, particularly when dealing with a large number of object categories. 
  Therefore, optimizing the model to reduce parameters and decrease model size is a crucial future research direction.
\end{itemize}

In summary, this study provides valuable insights into the application of bio-cognitive mechanisms to enhance computer vision algorithms, and future research can further advance the HB-net framework and explore its potential in various real-world scenarios.

\section*{Declaration of Competing Interest}

The authors declare that they have no known competing financial interests or personal relationships that could have appeared to influence the work reported in this paper.

\section*{Acknowledgments}

This work was supported in part by:
National Science Foundation of China under Grant 32171096, 31925018. 
Aeronautical Science Foundation of China, grant number 2020Z023053002.

\bibliography{plain}

\begin{thebibliography}{10}

\bibitem{Abbas2021}
Qaisar Abbas.
\newblock Frs-occ: Face recognition system for surveillance based on occlusion
  invariant technique.
\newblock {\em International Journal Of Computer Science And Network Security},
  21(8):288--296, AUG 30 2021.

\bibitem{Almalki2021}
Khalid~J. Almalki, Baek-Young Choi, Yu~Chen, and Sejun Song.
\newblock Characterizing scattered occlusions for effective dense-mode crowd
  counting.
\newblock In {\em 2021 IEEE/CVF International Conference On Computer Vision
  Workshops (ICCVW 2021)}, pages 3833--3842, 10662 Los Vaqueros Circle, Po Box
  3014, Los Alamitos, CA 90720-1264 USA, 2021. IEEE Computer Soc.

\bibitem{Cao2019}
Yanpeng Cao, Guizhong Fu, Jiangxin Yang, Yanlong Cao, and Michael~Ying Yang.
\newblock Accurate salient object detection via dense recurrent connections and
  residual-based hierarchical feature integration.
\newblock {\em Signal Processing-Image Communication}, 78:103--112, OCT 2019.

\bibitem{Cui2022}
Liyuan Cui, Zhiyuan Fan, Yingjian Yang, Rui Liu, Dajiang Wang, Yingying Feng,
  Jiahui Lu, and Yifeng Fan.
\newblock Deep learning in ischemic stroke imaging analysis: A comprehensive
  review.
\newblock {\em Biomed Research International}, 2022, NOV 14 2022.

\bibitem{Deng2020}
Xing Deng, Fepeng Da, Haijian Shao, and Yingtao Jiang.
\newblock A multi-scale three-dimensional face recognition approach with sparse
  representation-based classifier and fusion of local covariance descriptors.
\newblock {\em Computers \& Electrical Engineering}, 85:106700, JUL 2020.

\bibitem{Ding2020a}
Hui Ding, Peng Zhou, and Rama Chellappa.
\newblock Occlusion-adaptive deep network for robust facial expression
  recognition.
\newblock In {\em IEEE/IAPR International Joint Conference On Biometrics (IJCB
  2020)}, 345 E 47th St, New York, NY 10017 USA, 2020. IEEE; IAPR; IEEE Biometr
  Council; Google; Qualcomm; NSF, IEEE.

\bibitem{Ding2020}
Jisheng Ding, Linfeng Xu, Jiangtao Guo, and Shengxuan Dai.
\newblock Human detection in dense scene of classrooms.
\newblock In {\em 2020 IEEE International Conference On Image Processing
  (ICIP)}, pages 618--622, 345 E 47th St, New York, NY 10017 USA, 2020. Inst
  Elect \& Elect Engineers; Inst Elect \& Elect Engineers, Signal Proc Soc,
  IEEE.

\bibitem{Du2019}
Lingshuang Du and Haifeng Hu.
\newblock Nuclear norm based adapted occlusion dictionary learning for face
  recognition with occlusion and illumination changes.
\newblock {\em Neurocomputing}, 340:133--144, 2019.

\bibitem{Duan2021}
Qingyan Duan and Lei Zhang.
\newblock Look more into occlusion: Realistic face frontalization and
  recognition with boostgan.
\newblock {\em IEEE Transactions On Neural Networks And Learning Systems},
  32(1):214--228, JAN 2021.

\bibitem{Ernst2021}
Markus~R. Ernst, Thomas Burwick, and Jochen Triesch.
\newblock Recurrent processing improves occluded object recognition and gives
  rise to perceptual hysteresis.
\newblock {\em Journal Of Vision}, 21(13):6--6, DEC 2021.

\bibitem{Gao2021}
Jixun Gao and Yuanyuan Zhao.
\newblock Tfe: A transformer architecture for occlusion aware facial expression
  recognition.
\newblock {\em Frontiers In Neurorobotics}, 15:763100, OCT 25 2021.

\bibitem{Georgescu2022}
Mariana-Iuliana Georgescu, Georgian-Emilian Duta, and Radu~Tudor Ionescu.
\newblock Teacher-student training and triplet loss to reduce the effect of
  drastic face occlusion application to emotion recognition, gender
  identification and age estimation.
\newblock {\em Machine Vision And Applications}, 33(1):12, JAN 2022.

\bibitem{He2016}
Kaiming He, Xiangyu Zhang, Shaoqing Ren, and Jian Sun.
\newblock Deep residual learning for image recognition.
\newblock In {\em 2016 IEEE Conference On Computer Vision And Pattern
  Recognition (CVPR)}, pages 770--778, 345 E 47th St, New York, NY 10017 USA,
  2016. IEEE Comp Soc; Comp Vis Fdn, IEEE.

\bibitem{Kao2019}
ChiaoWen Kao, HuiHui Chen, BorJiunn Hwang, YuJu Huang, and KuoChin Fan.
\newblock Gender classification with jointing multiple models for occlusion
  images.
\newblock {\em Journal Of Information Science And Engineering}, 35(1):105--123,
  JAN 2019.

\bibitem{Kietzmann2019}
Tim~C. Kietzmann, Courtney~J. Spoerer, Lynn K.~A. Sörensen, Radoslaw~M. Cichy,
  Olaf Hauk, and Nikolaus Kriegeskorte.
\newblock Recurrence is required to capture the representational dynamics of
  the human visual system.
\newblock {\em Proceedings of the National Academy of Sciences},
  116(43):21854--21863, 2019.

\bibitem{Kortylewski2021}
Adam Kortylewski, Qing Liu, Angtian Wang, Yihong Sun, and Alan Yuille.
\newblock Compositional convolutional neural networks: A robust and
  interpretable model for object recognition under occlusion.
\newblock {\em International Journal Of Computer Vision}, 129(3):736--760, MAR
  2021.

\bibitem{Kortylewski2020}
Adam Kortylewski, Qing Liu, Huiyu Wang, Zhishuai Zhang, and Alan Yuille.
\newblock Combining compositional models and deep networks for robust object
  classification under occlusion.
\newblock In {\em 2020 IEEE Winter Conference On Applications Of Computer
  Vision (WACV)}, pages 1322--1330, 10662 Los Vaqueros Circle, Po Box 3014, Los
  Alamitos, CA 90720-1264 USA, 2020. IEEE Computer Soc.

\bibitem{Krizhevsky2017}
Alex Krizhevsky, Ilya Sutskever, and Geoffrey~E. Hinton.
\newblock Imagenet classification with deep convolutional neural networks.
\newblock {\em Communications Of The ACM}, 60(6):84--90, JUN 2017.

\bibitem{Li2020}
Kunjian Li and Qijun Zhao.
\newblock If-gan: Generative adversarial network for identity preserving facial
  image inpainting and frontalization.
\newblock In V~Struc and F~GomezFernandez, editors, {\em 2020 15th IEEE
  International Conference On Automatic Face And Gesture Recognition (FG
  2020)}, pages 45--52, 345 E 47th St, New York, NY 10017 USA, 2020. IEEE; IEEE
  Comp Soc; Univ Buenos Aires; Univ Palermo; 4Paradigm; Google; Wrnch AI;
  Empath Project, IEEE.

\bibitem{Li2021a}
Yande Li, Kun Guo, Yonggang Lu, and Li~Liu.
\newblock Cropping and attention based approach for masked face recognition.
\newblock {\em Applied Intelligence}, 51(5, SI):3012--3025, MAY 2021.

\bibitem{Li2022}
Yangyang Li, Zichen Yang, Yanqiao Chen, Danqing Yang, Ruijiao Liu, and Licheng
  Jiao.
\newblock Occluded person re-identification method based on multiscale features
  and human feature reconstruction.
\newblock {\em IEEE Access}, 10:98584--98592, 2022.

\bibitem{Liao2021}
Shangchun Liao, Gongfa Li, Hao Wu, Du~Jiang, Ying Liu, Juntong Yun, Yibo Liu,
  and Dalin Zhou.
\newblock Occlusion gesture recognition based on improvedssd.
\newblock {\em Concurrency And Computation-Practice \& Experience},
  33(6):e6063, MAR 25 2021.

\bibitem{Liu2023}
Chang Liu, Kaoru Hirota, and Yaping Dai.
\newblock Patch attention convolutional vision transformer for facial
  expression recognition with occlusion q.
\newblock {\em Information Sciences}, 619:781--794, JAN 2023.

\bibitem{Liu2022}
Zhuang Liu, Hanzi Mao, Chao-Yuan Wu, Christoph Feichtenhofer, Trevor Darrell,
  and Saining Xie.
\newblock A convnet for the 2020s.
\newblock In {\em 2022 IEEE/CVF Conference On Computer Vision And Pattern
  Recognition (CVPR)}, pages 11966--11976, 10662 Los Vaqueros Circle, Po Box
  3014, Los Alamitos, CA 90720-1264 USA, 2022. IEEE Computer Soc.

\bibitem{Lokku2022}
Gurukumar Lokku, G.~Harinatha Reddy, and M.~N.~Giri Prasad.
\newblock Opfacenet: Optimized face recognition network for noise and occlusion
  affected face images using hyperparameters tuned convolutional neural
  network.
\newblock {\em Applied Soft Computing}, 117:108365, MAR 2022.

\bibitem{Madarkar2020}
Jitendra Madarkar and Poonam Sharma.
\newblock Occluded face recognition using noncoherent dictionary.
\newblock {\em Journal Of Intelligent \& Fuzzy Systems}, 38(5):6423--6435,
  2020.

\bibitem{Peng2019}
Shenhui Peng, Sei-ichiro Kamata, and Toby~P. Breckon.
\newblock A ranking based attention approach for visual tracking.
\newblock In {\em 2019 IEEE International Conference On Image Processing
  (ICIP)}, pages 3073--3077, 345 E 47th St, New York, NY 10017 USA, 2019. IEEE.

\bibitem{Rajaei2019}
Karim Rajaei, Yalda Mohsenzadeh, Reza Ebrahimpour, and Seyed-Mahdi
  Khaligh-Razavi.
\newblock Beyond core object recognition: Recurrent processes account for
  object recognition under occlusion.
\newblock {\em PLOS Computational Biology}, 15(5):e1007001, MAY 2019.

\bibitem{Ran2021}
Xinyu Ran, Liang Ge, and Xiaofeng Zhang.
\newblock Rgan: Rethinking generative adversarial networks for cloud removal.
\newblock {\em International Journal Of Intelligent Systems},
  36(11):6731--6747, NOV 2021.

\bibitem{Sarapakdi2019}
Sittiphan Sarapakdi, Phaderm Nangsue, and Charnchai Pluempitiwiriyawej.
\newblock Occluded facial recognition with 2dpca based convolutional neural
  network.
\newblock In {\em 2019 4th IEEE International Conference On Consumer
  Electronics - Asia (IEEE ICCE-ASIA 2019)}, pages 135--138, 345 E 47th St, New
  York, NY 10017 USA, 2019. IEEE.

\bibitem{Shakeel2022}
M.~Saad Shakeel.
\newblock Bam: A bidirectional attention module for masked face recognition.
\newblock In {\em 2022 IEEE International Conference On Visual Communications
  And Image Processing (VCIP)}, 345 E 47th St, New York, NY 10017 USA, 2022.
  IEEE.

\bibitem{Sheng2023}
Biyun Sheng, Chaorun Sun, Fu~Xiao, Linqing Gui, and Zhengxin Guo.
\newblock Muat-va: Multi-attention and video-auxiliary network for device-free
  action recognition.
\newblock {\em IEEE Internet Of Things Journal}, 10(12):10870--10880, JUN 15
  2023.

\bibitem{Soni2021}
Neha Soni, Enakshi~Khular Sharma, and Amita Kapoor.
\newblock Hybrid meta-heuristic algorithm based deep neural network for face
  recognition.
\newblock {\em Journal Of Computational Science}, 51:101352, APR 2021.

\bibitem{Spoerer2020}
Courtney~J. Spoerer, Tim~C. Kietzmann, Johannes Mehrer, Ian Charest, and
  Nikolaus Kriegeskorte.
\newblock Recurrent neural networks can explain flexible trading of speed and
  accuracy in biological vision.
\newblock {\em Plos Computational Biology}, 16(10):e1008215, OCT 2020.

\bibitem{Spoerer2017}
Courtney~J. Spoerer, Patrick McClure, and Nikolaus Kriegeskorte.
\newblock Recurrent convolutional neural networks: A better model of biological
  object recognition.
\newblock {\em Frontiers In Psychology}, 8:1551, SEP 12 2017.

\bibitem{Sun2022}
Dengdi Sun, Wandong Xie, Zhuanlian Ding, and Jin Tang.
\newblock Silp-autoencoder for face de-occlusion.
\newblock {\em Neurocomputing}, 485:47--56, MAY 7 2022.

\bibitem{Wang2020}
Meng Wang, Xiang Liao, Ruijie Li, Shanshan Liang, Ran Ding, Jingcheng Li,
  Jianxiong Zhang, Wenjing He, Ke~Liu, Junxia Pan, Zhikai Zhao, Tong Li, Kuan
  Zhang, Xingyi Li, Jing Lyu, Zhenqiao Zhou, Zsuzsanna Varga, Yuanyuan Mi,
  Yi~Zhou, Junan Yan, Shaoqun Zeng, Jian~K. Liu, Arthur Konnerth, Israel
  Nelken, Hongbo Jia, and Xiaowei Chen.
\newblock Single-neuron representation of learned complex sounds in the
  auditory cortex.
\newblock {\em Nature Communications}, 11(1):4361, AUG 31 2020.

\bibitem{Xudong2019}
Xudong Wang, Hongquan Wei, Shaomei Li, Chao Gao, and Ruiyang Huang.
\newblock Unsupervised facial image occlusion detection with deep autoencoder.
\newblock In JN~Hwang and X~Jiang, editors, {\em Eleventh International
  Conference On Digital Image Processing (ICDIP 2019)}, volume 11179 of {\em
  Proceedings of SPIE}, 1000 20th St, Po Box 10, Bellingham, WA 98227-0010 USA,
  2019. E China Normal Univ; Int Assoc Comp Sci \& Informat Technol, Spie-Int
  Soc Optical Engineering.

\bibitem{Wang2019}
Yasi Wang, Hongxun Yao, Wei Yu, Dong Wang, Shangchen Zhou, and Xiaoshuai Sun.
\newblock Gradual recovery based occluded digit images recognition.
\newblock {\em Multimedia Tools And Applications}, 78(2):2571--2586, JAN 2019.

\bibitem{Wang2022}
Yawei Wang, Yifei Chen, and Dongfeng Wang.
\newblock Recognition of multi-modal fusion images with irregular interference.
\newblock {\em Peerj Computer Science}, 8:e1018, JUN 24 2022.

\bibitem{Wen2016}
Yandong Wen, Weiyang Liu, Meng Yang, Yuli Fu, Youjun Xiang, and Rui Hu.
\newblock Structured occlusion coding for robust face recognition.
\newblock {\em Neurocomputing}, 178(SI):11--24, FEB 20 2016.

\bibitem{Xia2023}
Xiaohan Xia and Dongmei Jiang.
\newblock Hit-mst: Dynamic facial expression recognition with hierarchical
  transformers and multi-scale spatiotemporal aggregation.
\newblock {\em Information Sciences}, 644:119301, OCT 2023.

\bibitem{Xiao2021}
Ma~Xiao, Chen Zhongwei, Suo Jun, Zhuansun Xiaobo, Ni~Jiazheng, Zhang Shuai, and
  Liu Mo.
\newblock A ship target recognition method based on biological visual attention
  mechanism.
\newblock In Y~Zhu and S~Xue, editors, {\em AOPC 2021: Novel Technologies And
  Instruments For Astronomical Multi-Band Observations}, volume 12069 of {\em
  Proceedings of SPIE}, 1000 20th St, Po Box 10, Bellingham, WA 98227-0010 USA,
  2021. Chinese Soc Opt Engn; Univ Elect Sci \& Technol China; Sci \& Technol
  Low Light Level Night Vis Lab; Sci \& Technol Electro Opt Informat Secur
  Control; Tsinghua Univ, Dept Elect Engn, Nano Optoelectron Lab, Spie-Int Soc
  Optical Engineering.

\bibitem{Yang2019}
Wanxiang Yang, Yan Yan, and Si~Chen.
\newblock Adaptive deep metric embeddings for person re-identification under
  occlusions.
\newblock {\em Neurocomputing}, 340:125--132, MAY 7 2019.

\bibitem{Yuan2016}
Li~Yuan, Wei Liu, and Yang Li.
\newblock Non-negative dictionary based sparse representation classification
  for ear recognition with occlusion.
\newblock {\em Neurocomputing}, 171:540--550, 2016.

\bibitem{Yuan2021}
Wenxue Yuan, Qijun Zhao, Feiyu Zhu, and Zhengxi Liu.
\newblock To see facial expressions through occlusions via adversarial
  disentangled features learning with 3d supervision.
\newblock In Jianjiang Feng, Junping Zhang, Manhua Liu, and Yuchun Fang,
  editors, {\em Biometric Recognition}, pages 84--91, Cham, 2021. Springer
  International Publishing.

\bibitem{Zheng2020}
Wenbo Zheng, Chao Gou, and FeiYue Wang.
\newblock A novel approach inspired by optic nerve characteristics for few-shot
  occluded face recognition.
\newblock {\em Neurocomputing}, 376:25--41, FEB 1 2020.

\bibitem{Zhu2018}
Jie Zhu, Fan Feng, and Bo~Shen.
\newblock People counting and pedestrian flow statistics based on convolutional
  neural network and recurrent neural network.
\newblock In {\em Proceedings 2018 33rd Youth Academic Annual Conference Of
  Chinese Association Of Automation (YAC)}, pages 993--998, 345 E 47th St, New
  York, NY 10017 USA, 2018. IEEE.

\bibitem{Zhu2021}
Wenqiu Zhu, Xiaoyi Wang, Yuezhong Wu, and Guang Zou.
\newblock A face occlusion removal and privacy protection method for iot
  devices based on generative adversarial networks.
\newblock {\em Wireless Communications \& Mobile Computing}, 2021:1--14, JUL 1
  2021.

\bibitem{Zou2020}
Tengtao Zou, Shangming Yang, Yun Zhang, and Mao Ye.
\newblock Attention guided neural network models for occluded pedestrian
  detection.
\newblock {\em Pattern Recognition Letters}, 131:91--97, MAR 2020.

\end{thebibliography}
\end{document}